\documentclass{llncs}

\usepackage{mmm}

\title
{Learning to Understand by Evolving Theories}

\author
{Martin E.~M\"uller and Madhura D.~Thosar}

\institute{Dept.~Comp.~Sci., Univ.~Appl.~Sciences Bonn-Rhein-Sieg, St.~Augustin, Germany\\
    Email: m.e.mueller@acm.org, madhura.thosar@gmail.com}

\newenvironment{madhucode}
{\\[.8ex] 
    \hspace*{.05\linewidth} \begin{minipage}{.9\linewidth}
      \small}
{\end{minipage}\\[.8ex] }

\begin{document}

\maketitle %

\begin{abstract}
In this paper, we describe an approach that enables an autonomous system to infer the
semantics of a \emph{command} (i.e.~a symbol sequence representing an action) in
terms of the relations between changes in the observations and the action instances.
We present a method of how to induce a theory (i.e.~a semantic description) of the
meaning of a \emph{command} in terms of a minimal  set of background knowledge.
The only thing we have is a sequence of observations from which we extract what kinds
of effects were caused by performing the command.
This way, we yield a description of the semantics of the action and, hence, a definition.
\end{abstract}

\section{Introduction}\label{sec:INTRODUCTION}

By ``semantics'' one usually refers to the meaning of something.
%
%
%
%
%
%
But since there is  no such thing as understanding in a computer, there is no
semantics either.
Yet, a robot \emph{acts} (causing events) and \emph{senses} the environment
(identifying things).
But  reasoning about actions requires  language and, hence, semantics.

There are several prime inspirations for our work: 
First,  John Locke's concept of learning from
scratch,
\cite{locke90:_essay_concer_human_under}, and
Bertrand Russell's ``Theory of knowledge'', \cite{Russell:1992aa}. \cite{Muller:2007aa} also
summarises several theories of cognition related to learning the semantics of an action.
%
%
%
With a model being  defined as a formal description of the observable effects of an action
we add to it the fundamental problem of identifying a change in the environment to be
an effect of an action. 
Both the environment and the innate knowledge is represented relationally as logic
programs so we can describe the formation of more complex models as a learning
process similar to inductive logic programming, \cite{Muller:2009aa}.

\section{Learning to Understand Actions}\label{sec:LEARNING}

Understanding and learning require knowledge. 
In this article, we presuppose an agent to have a minimal set of  ``knowledge'' as a given set of
terminological formulae describing facts, properties and rules that hold within the
agent's model of the world.


\subsection{Actions as state changes}\label{sec:LEARNING-SituationCalculus}


Performing an action usually results in a change in the world. 
Since we do not know the world itself but only a model of it, we need to live with
interpretations of  partial data collected by sensors.
Goal directed acting (planned, rational behaviour) therefore requires to anticipate
an action's results and we do so by assuming that if we \emph{do} something, the
real effects takes us into a situation where our sensors deliver results that
correspond to what we wanted to achieve:
\begin{equation}
  \begin{array}{c}
\xymatrix@C=6em@R=3ex{
  w_{i} \ar[r]^{\text{Action}} \ar[d]\ar[d]^{\val_i} & w_{i+1} \ar[d]^{\val_{i+1}} \\
\delta_{i} \ar[r]^{\text{Operator}} & \delta_{i+1}
}
\end{array}
\label{eq:actionsandoperators}
\end{equation}
All $\delta_i$ are \emph{state descriptions}, which consist of sets of logic formulae
(in our case, Horn clauses) over a signature $\Sigma$.
The $\val_i$ represent the interpretation of sensory data and, thus, the assignments
of the variables in $\delta_i$.
An operator $\mathit{op}$ is a function that changes $\delta_i$ and which anticipates
a change in $\val$.

\textbf{Example}. Let $\delta_i \models_{\val_i} \mathit{Position}=\tup{X,Y}$. Assume there is
a function $\mathit{mv}$ that takes as input the variable $\textit{Position}$ and two
further parameters $Dx$ and $Dy$ such that $\mathit{mv}(\tup{X,Y},  \mathit{Dx},
\mathit{Dy}) = \tup{X+\mathit{Dx}, Y+\mathit{Dy}}$. Then, we want $\delta_{i+1}
\models_{\val_{i+1}} \mathit{Position}=\tup{X,Y}$ where $\val_{i+1}(X) = \val_i(X) +
\mathit{Dx}$ and $\val_{i+1}(Y) = \val_i(Y) + \mathit{Dy}$.

Usually, the semantics of a formula $\phi$ is determined by an interpretation
function $\eval {} \val \cdot : \FML \to A$ with $A$ being the domain of the model
algebra.
For all variables and constant symbols $\eval {} \val X \defeq \val (X)$.
This means that in equation (\ref{eq:actionsandoperators}), $\val$ actually points the
other way round: The meaning of a variable is \emph{grounded} in the world.
The reason for us to denote the arrows the other way round is as follows:
We know the algorithmic (and thus purely syntactical) definition of $\mathit{op}$
(like $mv$) but we do not know its semantics.
So by observing the change $\mathtt{act}: \val_i \mapsto \val_{i+1}$ we find the
meaning of $\mathit{op}$ in a changed $\updelta$ such that
\begin{eqnarray}
  \label{eq:meaningbydeltaalpha}
  \val_{i+1} = \val_i \updelta 
  &\defeq&
  \val_i \nonumber
  \\ && - \set{ \tup{X, \val_i(X)} :
    \begin{array}{l}
      X \text{\ is affected while}\\
      \text{performing \texttt{act}}
    \end{array}}
  \\ && \cup
  \set{ \tup{X, \mathit{dX}} : 
    \begin{array}{l}
      X \text{\ is affected while}\\
      \text{performing \texttt{act} and its }\\
      \text{reading becomes\ } \mathit{dX}
    \end{array}\hspace*{-.8em}
}\nonumber
\end{eqnarray}
Note that ``being affected'' does not necessarily mean ``changed/caused by'':
Not everything that happens while we act is a consequence of our acting. Therefore,
observing a change in the environment does not mean it contributes to the meaning of
the performed action (c.f.~the frame problem and determinism,  \cite{McCarthy:1969aa}).
%
%
 
%

\subsection{Representing States}\label{sec:LEARNING-RepresentingStates}

We assume the world at a point $t$ in time to be represented by a \emph{state
  snapshot}.
State snapshots are, basically, total functions $\val_t$ which to a set of variables
representing certain sensory input assign corresponding values.
The set of variables is $\VAR = \VAR_O \cup \VAR_I$ with $\VAR_O = \set{\dot X_i : i \in
  \mathbf{n}_O}$ being the set of (global) observable variables and $\VAR_I = \set{X_j
  : j \in \mathbf{n}_I}$ being the set of (local) internal variables.
Furthermore, every variable $X \in \VAR$ is assigned a \emph{type}.
A state snapshot $s_t$ at time $t$ is defined as sequence
\begin{equation}
  \label{eq:statesnapshotformal}
  s_t \defeq \seq{\tup{X, \val_t(X)} : X \in \VAR}.
\end{equation}
For a set $T=\set{i : i \in \mathbf{m}}$ we have a \emph{sample}
\begin{equation}
  \label{eq:samplesnapshots}
  \mathbf{S} = \set{ \tup{t, s_t} : t \in T}
\end{equation}
of state snapshots.
Since $T$ is a strict order, we can also uniquely arrange all state snapshots in a
table which gives every variable $\dot X$ the reading of a function (i.e.~a column in
the table). Using the relational database definition language, the sample can be
described as 
\begin{equation}
\begin{array}{l}
  \mathrm{snapshots}(t, f_{\dot X_0}, \ldots, f_{\dot X_{n_O-1}},
  \val_t(X_0),  \ldots, \val(X_{n_O-1})) \\
  \mathrm{PK}(\mathrm{snapshots})=t\\
  \dom {f_{\dot X_i}} = d_i, \dom{\val(X_j)} = d_j
\end{array}\label{eq:statesnapshotRDB}
\end{equation}
where all $d_i, d_j$ are the types of the according variables and $t$ is a unique
identifier (``Primary Key'').\footnote{Therefore, equations
  (\ref{eq:statesnapshotformal}) and (\ref{eq:statesnapshotRDB}) are equivalent:
  $\mathrm{snapshot}(t, \ldots) = s_t$.}
We shall represent collected state snapshots as factual knowledge represented using simple
Horn logic facts and key-value lists:
\begin{equation}
  \label{eq:statesnapshotprolog}
  \mathrm{state}(t, \seq{ x_\mathrm{0}\!\!-\!\!\mathrm{v}_\mathrm{0}, x_\mathrm{1}\!\!-\!\!\mathrm{v}_\mathrm{1}, \ldots, x_{n-1}\!\!-\!\!\mathrm{v}_{n-1} }).
\end{equation}
where all $x_{\mathrm{i}}$ are variable \emph{names} (hence lowercase; they correspond
to the column features in the database notation in equation (\ref{eq:statesnapshotRDB})) and all
$\mathrm{v}_i$ are the \emph{instantiations} of the according variables at time $t$.
%
Table \ref{tab:variabletypes-5.1:5.2} lists the hierarchy of variable types.
%
%

\subsection{Bootstrapping initial theories}\label{sec:LEARNING-Bootstraping}

Symbol grounding or lexical semantics fails to work until one agrees on a set of
common, atomic meaningful symbols from which one can construct larger ones.
Whatever we perceive, we need to be able to put it into words in order to reason
about it. 
And whatever we shall learn can be learned only if we have a sufficient repository of
basic categories from which we can build a desired concept.
%
Our initial knowledge consists of a type system and a set of primitive typed relations
and (arithmetic) operations.
It contains the definitions of relations such as greater than, less than; arithmetic
relations such as addition, subtraction, multiplication, division; spatial relations
like  position, orientation, proximity; and predicate logic expressions including
equality.
The type system can easily be adapted to the domain:

\textbf{Example.} With only real numbers $\mathtt{num}:X$ and $\mathtt{num}:Y$ describing
coordinates  and suitable arithmetic operations  we can define a type $\mathtt{pos}$ as
$\mathtt{num}\times\mathtt{num}$ with a new operation
$\mathit{dist}:\mathtt{pos}\to\mathtt{pos}$ as
\begin{table}[htbp]
  \centering
  \caption{Variable types}
  \label{tab:variabletypes-5.1:5.2}
  \begin{tabular}{llll}
    \hline No.& Type & Example & Interpretation \\ \hline
    1 & \texttt{num} & \texttt{num}:5 & Cardinals and ordinals \\
    2 & \texttt{pos} & \texttt{pos}:$\tup{x,y,x}$ & Positioning (complex \texttt{num}
    )\\
    3 & \texttt{dist}& \texttt{dist}:5 & \texttt{num} denoting distance \\
    4 & \texttt{angl}& \texttt{angl}:90 & \texttt{num} $0 \leq x < 360$, angular
    orientation\\
    5 & \texttt{bool}& \texttt{bool}:1 & Truth value \\
    6 & \texttt{obj}& \texttt{obj}:ball & Generic object identifier \\ \hline
    \multicolumn{3}{c}{Class} & Subsumed types \\ \hline
     \multicolumn{3}{c}{\texttt{arith}} & \texttt{num}, \texttt{pos}, \texttt{dist} \\
     \multicolumn{3}{c}{\texttt{comp}} & \texttt{num}, \texttt{obj} \\
     \multicolumn{3}{c}{\texttt{spatial}} & \texttt{pos}, \texttt{dist}, \texttt{angl}, \texttt{obj} \\
     \multicolumn{3}{c}{\texttt{logic}} & \texttt{bool}, \texttt{object}
  \end{tabular}
\end{table}
\[
\mathit{dist} ( 
   P_1, P_2 ) 
   \defeq \mathit{sqrt}( (P_1.1-P_2.1)\text{\texttt{\^}}2 \mathtt{+} (P_1.2-P_2.2)\text{\texttt{\^}}2 )
\] 
where $X.i$ denotes the $i$-th component of $X$.
This way, the background knowledge can be represented as follows:

\textbf{Example.} First, we define the signature of the operations or relations to be
implemented: 
\begin{eqnarray*}
  \mathit{add\_to} &:& \mathtt{num} \times \mathtt{num} \to \mathtt{num} \\
  \mathit{left\_of} &:& \mathtt{obj} \times \mathtt{obj}  \to \set{\mathrm{true}, \mathrm{false}}
\end{eqnarray*}
The according definitions are  straightforward:
\[
\begin{array}{lcl}
  \mathrm{add\_to}(\mathtt{num}:X, \mathtt{num}:Y, \mathtt{num}\mathop :Z)  
  &\defEqu& Z = X +Y.\\
  \mathrm{left\_of}(\mathtt{pos}:(X1,\_), \mathtt{pos}:(X2,\_)) 
  &\defEqu& X1 < X2.
\end{array}
\]
%

\subsection{Actions}\label{sec:LEARNING-Actions}

An \emph{action} is something a robot does. 
Usually, every action is the result of a \emph{command} and it causes several effects
(hopefully all and only desired).
A command is an instantiation of an \emph{operator} (c.f.~STRIPS,
\cite{Nilsson:1970aa}):
\[
  \label{eq:STRIPS-Op-Move}
  \left[
    \begin{array}{ll}
      \mathtt{MOVE} & \mathtt{obj}:\mathit{Obj}, \mathtt{num}:x \\
      \mathtt{PRE}  & \set{\mathit{at}(\mathit{Obj}, \mathtt{pos}:\tup{\dot X, \dot
          Y})} \\
      \mathtt{EFF} &  
      \seq{
      \begin{array}{l}
        \mathtt{power}(0.8); 
        \mathtt{wait}(c \cdot x); 
        \mathtt{power}(0)        
      \end{array}
      }\\
      \mathtt{DEL} &  \set{\mathit{at}(\mathit{Obj}, \mathtt{pos}:\tup{\dot X, \dot
          Y})} \\
      \mathtt{ADD} &  \set{\mathit{at}(\mathit{Obj}, \mathtt{pos}:\tup{
          \dot X + c x  \sin \dot Z, 
          \dot Y + c x \cos \dot Z
         })} \\
    \end{array}
  \right]
\]
One might state that the semantics of moving is rather simple: it is the program as
it is defined in the \texttt{EFF}-slot of the operator definition, but our goal is to
explain, what it \emph{means} to do \texttt{EFF}.


Note the different kinds of variables used in the operator definition above: As
already stated, the dotted variables are global variables holding feature values that
describe the world.
Simple variable symbols are required for an abstract definition of the operator.
\texttt{MOVE} works for any kind of object $\mathit{Obj}$, accepts as a parameter 
a numerical value $x$ (which means the distance one shall move), and a constant $c$ 
that translates distance into the period of time one travels at a determined speed.

In planning, one puts the semantics into the operator definition by defining changes
in the state descriptions.
The operator is applicable only if $\texttt{PRE}$ is satisfied and $\mathtt{ADD}$ and
$\mathtt{DEL}$ describe how a state description changes after
performing $\mathtt{EFF}$.\footnote{Note that in this paradigm, one works on
  \emph{descriptions} of states. In a closed, deterministic and fully observed world such a
  description (is assumed to) coincide with reality.
  In the real world, however, situations may change \emph{without} a change of its description
  and there might be different description for one and the same world.}
As one can see, the new position of the robot after moving is defined in terms of the
previous position ($\dot X$ and $\dot Y$), the travel distance $x$ (and a factor $c$)
and another globally observable feature $\dot Z$: the robot's orientation.

\subsection{Learning by Observing}\label{sec:LEARNING-Observing}

If we want a robot to learn the semantics of its actions, we reverse the line of
argument:
$\mathtt{MOVE}$ means that \emph{certain} variables change in a \emph{certain} way.
This, in general, gives rise to two learning problems:
\begin{enumerate}
\item Which is the (smallest) subset $V \subseteq \VAR$ that contains all relevant
  variables for describing the effects of an action?
\item Once we know where to look for changes, what is the underlying system in the
  way the values change?
\end{enumerate}
The first questions seems simple to be answered: Collect all those variables $\dot
X$ with $\val_i(\dot X) \neq \val_{i+1}(\dot X)$ for an action that is performed
between $i$ and $i+1$.
But this is too simple: First, the meaning of an action could be exactly to
\emph{preserve} a certain value rather than changing it and second there are many
other reasons why a variable might be changed without any connection to performing
the action.
The second question is not as simple.
Let $V \subseteq \VAR_O$ be a suitable set of observable variables. 
Then, the question is to find the general law that describes the change. 
This presupposes that operators are ``synchronised'' in the sense, that all affected
variables are changed at the same time.\footnote{A first step towards dealing with
  state snapshots of asynchronous processes in the context of learning was discussed
  in \cite{Muller:2008ab}.}
With a suitable set of background knowledge (see sections
\ref{sec:LEARNING-RepresentingStates}-~\ref{sec:LEARNING-Actions}) and Horn logic as
representation formalisms, Inductive Logic Programming (ILP),
\cite{Bun:87,mug:cigol1,mugg:der,Raedt:2008aa},  is a prime candidate for 
learning the semantics of action.  

\subsection{Related work}

DIDO, \cite{Scott89learningnovel}, explores  unfamiliar domains without any external
supervision.
The task is to build a representation of the domain which can be used to predict the outcomes 
of its motor operations with a maximum likelihood of being correct.
Its key features - unsupervised inductive learning from actions and state
descriptions without prior knowledge - represent the fundamental principles on which
EVOLT is based on, too.
The major difference between the proposed system and DIDO is that EVOLT learns by
observing the environment whereas DIDO learns by the (given) effects of an action.

LIVE, \cite{Shen92complementarydiscrimination,Shen94discoveryas}, uses \emph{complementary discrimination learning} which is inspired by
Piaget's child development theories to learn the prediction models of effects of an
action.  
A prediction model in this setting is given by the triplet
\emph{$\langle$precondition, action, prediction$\rangle$} and  states that if an
action applied to a percept which satisfies a precondition then the resulting percept
should satisfy the prediction.  

Like DIDO, HYPER \cite{Leban:2008:ERD:1431080.1431092} shall learn  by exploring the domain. 
HYPER uses standard ILP (Aleph, \cite{Aleph})  to induce knowledge about movement (movability, obstacle
identification, DoF).
In the course of the XPERO project \cite{kahl09:_xpero_learn_exper_d1} the same ILP learner  was used. 
One one hand, this requires a far more elaborate background knowledge which in turn
introduces bias in form of model assumptions. On the other hand, full ILP
requires negative examples to avoid over-generalisation.
\cite{akhtar11:_towar_ilp} focus on generating negative examples by introducing a
bias that determines feature value ranges.

While we deal with more or less the same problem of learning the semantics of
actions, our approach differs from all these approaches  as we try to learn without negative
examples from scratch or, at least, with as few as possible assumptions
and predetermined (and predetermining) knowledge.

\section{EVOLT: Evolving Theories}\label{sec:EVOLT}

Learning the meaning of an action in EVOLT comprises of three stages of subsequent
steps:\footnote{Here, we use the term ``theory'' for what one usually calls a ``model''.}
\begin{enumerate}
\item Identify relevant observable variables.
\item Induce a new hypothesis describing the performed action.
\item Refine already existing hypotheses.
\end{enumerate}
%
%
In this first study and to suit the huge amount of flat data we have focused on one
aspect of ILP and extended the generalisation procedure to deal with typed terms.
%
%
Additionally, 
simple data collections leave us with 
positive examples only.
Finally, the clause head of the target predicate is under-specified in terms of mode
description as, e.g.~used \cite{muggsrin:progol} such that there  are no prior
candidates for the sought clause head available.

EVOLT searches for the relations exhaustively  where each
relation definition in background knowledge is scanned to check whether it entails
the changes in the state variables caused by an action.
%
%
The increase of search effort due to the brute-force method is alleviated by our use
of a type system (cf.~\ref{sec:LEARNING-RepresentingStates}) which allows quick
unification test.
Nevertheless, this efficiency gain does not scale for the number of subsets of
variables still grows exponentially with a linear growth in the number of variables
used on a base determined by the size of a variables domain.

In the following, different stages of learning are described briefly.
\subsection{Evolving theories by stepwise refinement}\label{sec:EVOLT-Evolving}

\subsubsection{Extracting action effects from  state description data}
\label{sec:EVOLT-Evolving-extracting}

We shall explain this by the following example:
\begin{figure}[!h]
  \centering
   \includegraphics[width=8.0cm]{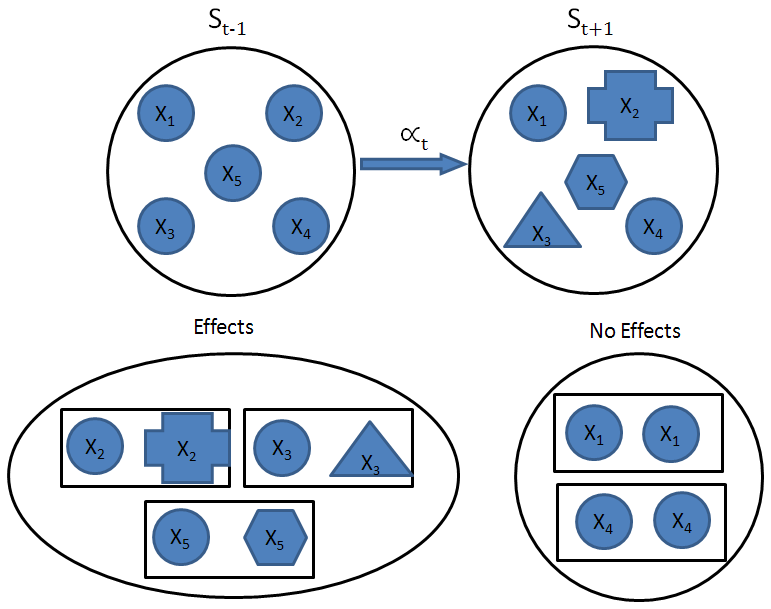}
  \caption{Transforming the set of state description associated with an action into a
    set of state transition.} 
  \label{fig:transform}
 \end{figure} 
 Each shape\footnote{Similar shapes in the same set do not necessarily represent the
   same value for corresponding variables. However similar shapes corresponding to
   the same variable in different sets represent the same value.} in figure
 \ref{fig:transform} represents a value assigned to a variable $\dot X_{i}$.  
Assuming that at time $t$ an action was performed and the immediate prior and
posterior state snapshots are $s_{t-1}$ and $s_{t+1}$, we define an \emph{effect}
and a \emph{no-effect set}
\begin{eqnarray}
  \label{eq:1}
  \mathit{Eff}_t &\defeq& \set{X \in \VAR_O : \val_{t-1}(X) \neq \val_{t+1}(X)} \\
  \comp{\mathit{Eff}_t} & \defeq & \VAR_O - \mathit{Eff}_t
\end{eqnarray}
respectively.
%
Imagine now that at $t$ (that is, between $t-1$ and $t+1$) a command to perform
$\mathtt{act}$ had been issued and all effects (``\texttt{EFF}'') came into force.
Assume further that $\mathtt{act}$ was called by parameters $a_0, \ldots, a_{k-1}$.
Therefore we expect to see that $\mathtt{act}_t( a_0, \ldots, a_{k-1})$ causes a
certain change:
\[
\mathtt{act}_t( a_0, \ldots, a_{k-1}) : \val_{t-1} \mapsto \val_{t+1}
\]
where $\val_{t+1} = \val_{t-1} \subst$ and $\subst$ is the substitution acting on
$\val_{t-1}$ as defined by the operator declaration that was executed.
In analogy to our example in section \ref{sec:LEARNING-Actions}, we have the
following 

\textbf{Example.} Let $Obj = \mathrm{'huey'}$ and $x=45$. Assume $\dot X = 0$ and $\dot Y =
7$. Then (presupposing a suitable value $c$), we would expect the robot Huey to end
up on a position that is ``northeast'' of its initial position:
\begin{eqnarray*}
\mathtt{MOVE}_t('\mathrm{huey}', 45) (\dot X) &:& 0 \mapsto 5 \\
\mathtt{MOVE}_t('\mathrm{huey}', 45) (\dot Y) &:& 7 \mapsto 12 
\end{eqnarray*}

\subsubsection{Building hypothesis about cause and effect}
\label{sec:EVOLT-Evolving-building}

Hypotheses are formulated in terms of possible linkings between action parameters
$a_i$ and variables $\dot X$.

\begin{figure}[!h]
  \centering
   \includegraphics[width=8.0cm]{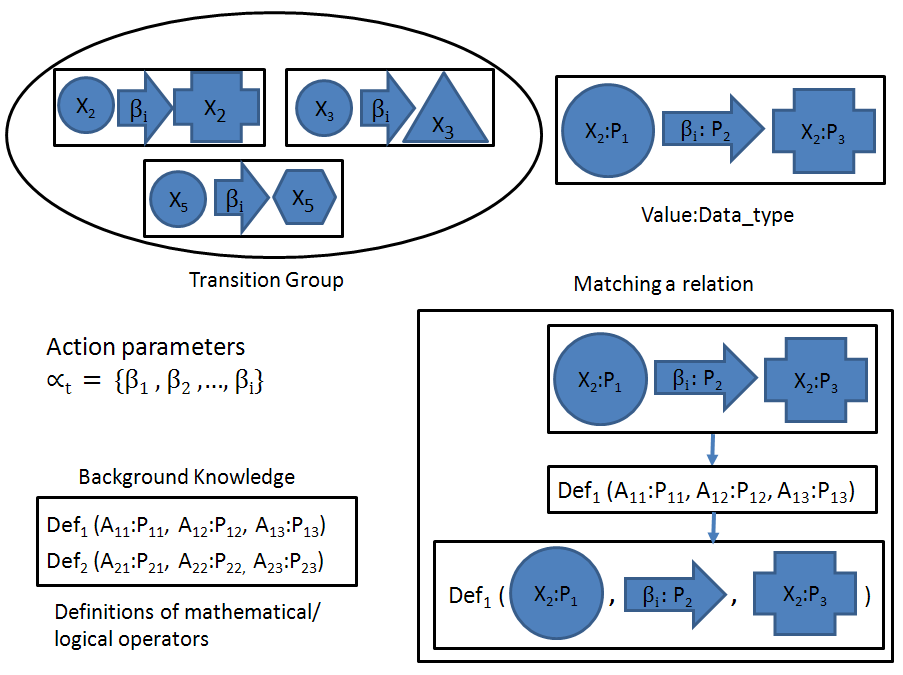}
  \caption{Learning a new theory.}
  \label{fig:learn}
 \end{figure}
The transition group $\mathit{Tran}$ is set of triplets containing all combinations
of observable variables and parameters:
\begin{equation}
  \label{eq:TransitionGroup}
  \mathit{Tran}_t \defeq \set{\tup{\val_{t-1}(\dot X), a_i, \val_{t+1}(\dot X)} : 
    \begin{array}{l}
      \dot X \in \mathit{Eff}_t, \\ 0 \leq i < k
  \end{array}
  }
\end{equation}
%
If we assume there are $m$ effect pairs, then for $k$  action
parameters, there will be $k \cdot m$ number of triplets in a transition group. 
Since we examine only one action at a time, $k$ is restricted by the number of
parameters of only one operator which, in general, is rather small. 
As shown in figure \ref{fig:learn}, each member of the triplet is assigned a data type (recall
section \ref{tab:variabletypes-5.1:5.2}) which acts as a search bias to reduce the
search space.  
The background knowledge provides the  definitions of predicates that can be  used by
the learner to explain the relation which may exist between instantiations of
variables $\dot X_{j}$ caused by an action $\mathtt{act}$ 
with parameters $a_i$. 
Once the transition group is created, the learner examines each triplet against each
definition in the background knowledge to identify a relation described by values in
a given triplet by matching the predicate signature against the types in the
transition group:
\begin{equation}
  \label{eq:BuildingAnInitialTheory}
  \mathit{Th}^\mathtt{act}_t 
  \defeq
  \set{
    \tup{\dot X, R} : 
    \begin{array}{l}
      \tup{\val_{t-1}(\dot X), a_i, \val_{t+1}(\dot X)} \in \mathit{Tran}_t,
      \\
      R = \set{r \;:\; r\!:\!s_1 \times s_2 \to s_3}
    \end{array}
    }
\end{equation}
where the signature matches the types of the transition triplet.

\subsubsection{Refining  existing theories} \label{sec:EVOLT-Evolving-RefinementStepIII}

%

Assume we already inferred a theory $\mathit{Th}_t^\mathtt{act}$.
Now, at $t'>t$, the transition group suggests to build a new theory
$\mathit{Th}_{t'}^\mathtt{act}$.
In order to refine  $\mathit{Th}_t^\mathtt{act}$ to meet the hypotheses of
$\mathit{Th}_{t'}^\mathtt{act}$ we build the intersection which then contains only
those relation/variable candidates that are part of both theories.
This way, we monotonously specify the theory by ruling out all those candidates that
do not support all of our observations:
\begin{equation}
\mathit{Th}_{t+1}^\mathtt{act} \defeq \mathit{Th}_t^\mathtt{act} \cap
\mathit{Th}_{t-1}^\mathtt{act}.\label{eq:theory-update} 
\end{equation}
While an on-line learning method that continuously refines a theory is a very useful
tool for robots that learn from scratch, the brute intersection has disadvantages, too:
\begin{itemize}
\item It could be an operator does not \emph{always} affect the same set of variables.
  For example, one could move into the $y$-direction only thereby leaving the
  $x$-coordinates unchanged. But a \texttt{move} usually affects \emph{all}
  components of the location.
\item Forcing every affected variable to be affected every time is also prone to
  noise since it could be we just have an observation or quantisation error.
\end{itemize}
The motivation for choosing this approach despite its disadvantages was that in real
life the non-determinism of the surrounding world results in a \emph{huge} set of
transitions pairs, and, hence, in large theories. 
Therefore we rule out singular observations to decrease noise by the price of over-pruning
the hypothesis space.
The results described in the next sections are surprisingly accurate and semantically
valid and therefore support our approach to theory refinement by monotonic specialisation.

\subsection{Example}\label{sec:EVOLT-Example}

\subsubsection{State descriptions}\label{sec:EVOLT-Example-state-desciptions}

The following is an example of the EVOLT representation of states (section \ref{sec:LEARNING-RepresentingStates}).
Each state description is represented by a predicate $\mathtt{state\_spec/2}$ with
its first argument being the time stamp $t$ and the second one a variable list
(c.f.~equation (\ref{eq:statesnapshotprolog})).
For example, 
\begin{madhucode}
  \begin{tabbing}
  state\_spec(31, [ \\
  \ \= [r\_pos,\ \ \ \ \ \ \= 
          [9, 14]{:pos]},\ \ \ \=\\
    \> [{obj\_num,} 
      \> {1:num}],\\
  \> [{obj\_grab}, 
     \> [{none}]{:obj},
     \> {0:truthVal}],\\
  \> [{obj\_pos},
     \> [{obst}]{:obj},
     \> [13,3]{:pos}]\\
     \> ]).
\end{tabbing}
\end{madhucode}
Each variable is, again, represented as a list with the first list element being the
name of an observable variable $\dot X$ (e.g.~$\mathtt{obj\_con}$) and the remaining
arguments being its value $\val_t(\dot X)$.
By unification, $[\dot X | \mathit{Val}]$, $\mathit{Val}$ is a list of typed variable values (with types written
postfix). 
This allows to assign complex values to a variable: $ [\mathrm{obj\_pos},
  [\mathrm{obst}]{:\mathtt{obj}}, [{13,3}]{:pos}]$ means that $\dot{Obj\_Loc}$ is of a type
$\mathtt{obj} \times \mathtt{pos}$. 
The state snapshot states that at time $31$ the robot is at $\tup{9,14}$, there is one more object, an
obstacle is at position $\tup{13,3}$, and the robot does not hold anything in its
gripper.

\subsubsection{Actions}
Actions are represented  by a predicate $\mathtt{action}(\mathit{Action\_id}, \mathit{Time},
  \mathit{Para}$- $\mathit{meters})$.
The first argument serves as an identifier for program internal purposes, 
$\mathit{Time}$ identifies the time stamp $t$ and $\mathit{Parameters}$ is a list of
parameters.
For example, the action \emph{move forward} by a distance $3$ at time $32$ is
represented as
\begin{madhucode}
  action(move\_forward, 32, [3:dist]).
\end{madhucode}

\subsubsection{Action-effect Theory}

An \emph{action theory} $\mathit{Th}^\mathtt{act}_t$ is represented by
$\mathtt{action\_theory/2}$.
Since we continuously update theories (see equation (\ref{eq:theory-update}), we do
not explicitly store the time stamp.
However, multiple alternative hypotheses can be asserted to the Prolog workspace thus
enabling a time-independent off-line learning on entire sets of theories.
\begin{madhucode}
  action\_theory(theory(Name), Params, relation\_is(Relations))
\end{madhucode}
The first argument carries the name $\mathtt{act}$ of the action to be explained, the
second argument is a list of parameters that were passed to $\mathtt{act}$.
$\mathtt{relation\_is}(\mathit{Relations})$ holds pairs of observable variables $\dot X$ and the
candidates for operators/relations that can be used to describe parameter-dependent
changes (see equation (\ref{eq:BuildingAnInitialTheory})). 

Again, we give an example for moving forward:
\begin{madhucode}
  \begin{tabbing}
  act\=ion\_theory(\\
     \> \texttt{move\_forward}, \\
     \> [$D$:dist], \\
     \> rel\=ation\_is([ \\ 
     \> \ [ r\_pos, \\
     \> \> [ has\=\_new\_position( \\
     \> \> \>  [$X_1,Y_1$]:\=pos, \\
     \> \> \>  $D$:\>dist, \\
     \> \> \>  [$X_2,Y_2$]:\>pos)\\
\> \> ]]])\\
).
\end{tabbing}
\end{madhucode}
It states the  trivial but true proposition that, when moving, the
position changes.

\subsubsection{Background Knowledge}

As seen in the previous example, the $\mathit{Th}^\mathtt{act}_t$ consists of
prior defined relations and operators.
Examining the background knowledge for relations satisfying the  changes as recorded
in $\mathit{Tran}_t$ involves two steps.  
In the first step, the data type of the triplet is matched against the data type of 
the operator declarations. 
This is a simple unification problem for the applicability of respective predicates.
The background knowledge predicate definitions for the example from section
\ref{sec:LEARNING-Bootstraping} are:
\begin{madhucode}
  \begin{tabbing}
    add\_to \= $(X:\mathtt{num}, Y:\mathtt{num}, Z:\mathtt{num})  \pif$ \\
            \> $Z = X +Y.$\\
    left\_of ($[X_1,\_]:\mathtt{pos}, [X_2,\_]:\mathtt{pos}) \pif$ \\ 
            \> $X_1 < X_2$.
  \end{tabbing}
\end{madhucode}
\textbf{Example.} 
The idea is that, based on the state description for $t=31$ and an according state
description at $t=33$ including [r\_pos, [9, 20]{:pos]}], and with background knowledge
\begin{madhucode}
  \begin{tabbing}
    travel\_x\=($[X_1,Y]:\mathtt{pos}, D:\mathtt{dist}, [X_2,Y]:\mathtt{pos}) \pif$ \\ 
            \> orientation\_WE($Z$)\\
            \> $X_2$ is ($X_1$ + ($Z$ * $C$ * $D$)).
  \end{tabbing}
\end{madhucode}
the system deduces that carrying out a task with name $\mathtt{move\_forward}$ at $t=32$ and an
action parameter $3$ when heading East (then, $Z=1$; heading West means $Z=-1$)
results in an $x$-aligned position change:
\begin{madhucode}
  \begin{tabbing}
  act\=ion\_theory(\\
     \> \texttt{move\_forward}, \\
     \> [$D$:dist], \\
     \> [ travel\=\_x(\\
     \> \>  [$X_1,Y_1$]:\=pos, \\
     \> \>  $D$:\>dist, \\
     \> \>  [$X_2,Y_1$]:\>pos)\\
     \> \> ]\\
     \>).
\end{tabbing}
\end{madhucode}
because the above goal is provable with instantiating $C=2$.


\subsection{Results}\label{sec:EVOLT-Results}
\begin{figure}[!h]
  \centering
   \includegraphics[width=5.5cm]{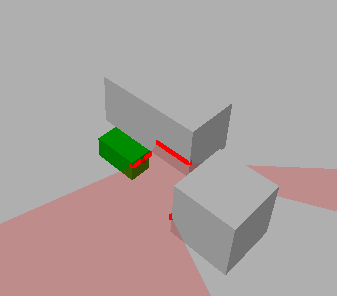}
  \caption{Task environment}
  \label{fig:taskenvironment}
 \end{figure}

%
The primary goal of the experimental evaluation was to examine the applicability and
behaviour of EVOLT in different environmental settings.   
The example of one such environmental setting is given below.
%
%
The data (set of states and actions) was collected as follows:
A robot was controlled to move in a simulated environment (using openRAVE). We
recorded all state descriptions (by reading out the simulation system variables and
representing them as facts) and also recorded all the commands that were issued to
the robot (that is, the manual control commands were translated into action
commands).
Figure \ref{fig:taskenvironment}  depicts the task environment that was considered
for an experiment where the robot is placed between two obstacles.
In our case, the position is expressed using a three-dimensional coordinate system
(adding height to our previous code examples).
We tried learning semantic descriptions of several primitive actions like \emph{move
  forward}, \emph{turn left/right}, \emph{grab/drop object} based on background knowledge.
The following is an example of the semantics learned for \emph{turn left/right}
(disregarding all information about the two obstacles):
\begin{madhucode}
  \begin{tabbing}
     r\_p\=os, \\
         \> [ cha\= nge\_in\_orientation(\\
         \>      \> [$X$, $Y$] \= :pos,\\
         \>      \>  $G$       \> :angle,\\ 
         \>      \> [$X$, $Y$] \> :pos )\\
         \> ]
  \end{tabbing}
\end{madhucode}
meaning that 
whenever an action name ``$\mathtt{turn}\__X$'' is performed, and whenever it receives
a parameter $G$ (which is an angle), then the orientation of the robot changes, but
its position remains the same. 

\section{Conclusion}\label{sec:CONCLUSION}

\subsection{Summary}\label{sec:CONCLUSION-Summary}

%
We have proposed an approach to learn the semantics of actions by describing the
observed effects in terms of atomic arithmetic/logical operations.
An action is considered as an event that causes the current state to change where the
state
is a set of features which represent sensory inputs. 
The working principles of EVOLT are inspired by the Evolution of Theories paradigm, 
while the learning itself is implemented using principles of  Inductive Logic Programming.
EVOLT uses exhaustive or brute-force search to search the relations where the search
is guided by the search bias in the form of typed variables.
Another contribution is that, motivated by its use in robotics, our systems works on
unlabelled examples.
Moreover, EVOLT is provided with a minimal set of background knowledge which is
``common knowledge'' 
in a sense that it is not task specific and only provides atomic operations on
sensory data.
Even though the results of the evaluation are quite reasonable, the system is still
in its adolescent stage.   
For example, it is assumed that all actions are executed successfully, that there are
no side-effects and that the world is fully observable and deterministic. 
However, all  of these requirements usually are not met in a real world robot
scenario.

\subsection{Prospects}

Our immediate future goal is to make the system more sophisticated by improving its 
data handling and extending background knowledge with additional definitions of more
complex relations. 
Further space of improvement is given by needs for
\begin{itemize}
\item a type system that allows typing of complex terms,
\item added  time stamps to action-theory facts for sequence learning,
\item building  transition sets on entire $\mathit{Eff}_t \times \mathbf{k} \times
  \mathit{Eff}_t$, therefore allowing to express semantic relations involving
  \emph{several} variables,
\item considering as transitions $\powerset{\mathit{Eff}_t} \times \powerset{\mathbf{k}}
  \times \powerset{\mathit{Eff}_t}$ which takes the learning to a further level of
  abstraction by combining \emph{several} observable values,
\item a more sophisticated theory refinement that goes beyond a mere collection of
  common appearances of variable assignment changes.
\end{itemize}
With a then growing complexity of search space we also need to introduce more
sophisticated biases.
Also, the dimensionality of hypothesis space (as determined by the number of
observable variables) needs to be \emph{reduced}, where at the same time we want to enable
the system to deal with  an \emph{increasing} number of features.
One method for a feature selection procedure in the context of relational
representation of data is rough set data analysis, \cite{Muller:2009aa}; see also
sections \ref{sec:LEARNING-RepresentingStates} and \ref{sec:LEARNING-Observing}).
Together with a modal logic representation form we aim at building a framework for a
modal inductive logic programming setting that shall be able to infer more powerful
semantic descriptions of actions.

{\footnotesize
\emph{Thanks.} The authors wish to thank Bj\"orn Kahl for his support and expertise
in robot manipulation and simulation.
}


\bibliographystyle{splncs}

\bibliography{mmm}

\begin{thebibliography}{10}

\bibitem{locke90:_essay_concer_human_under}
Locke, J.:
\newblock An essay concerning human understanding.
\newblock The Pennsylvania State University (1690)

\bibitem{Russell:1992aa}
Russell, B.A.W.:
\newblock Theory of Knowledge: The 1913 Manuscript.
\newblock Routledge (1992)

\bibitem{Muller:2007aa}
M{\"u}ller, M.E.:
\newblock {Being aware: Where we think the Action is}.
\newblock {Cognition, Technology, and Work} \textbf{9}(2) (2007)  109--126

\bibitem{kahl09:_xpero_learn_exper_d1}
Kahl, B., Henne, T., K{\"o}ckemann, U., Shahazad, C., Prassler, E.:
\newblock Xpero - learning by experimentation (d1.6).
\newblock Workpackage final report, German Natl.~Science Foundation (October
  2009)

\bibitem{akhtar11:_towar_ilp}
Akhtar, N., Fueller, M., Kahl, B., Henne, T.:
\newblock Towards iterative learning of autonomous robots using ilp.
\newblock In: 15th Intl.~Conf.~on Advanced Robotics (ICAR). (2011)  409--414

\bibitem{Muller:2009aa}
M{\"u}ller, M.E.:
\newblock Relational Knowledge Discovery.
\newblock Cambridge University Press (2012)

\bibitem{mccarthy:1959}
McCarthy, J.:
\newblock Programs with commonsense.
\newblock In: Proceedings of the {T}eddington Conference on the Mechanization
  of Thought Processes.
\newblock Her Majesty's Stationery Office, London (1959)  75--91 Reprinted
  (with an added section on `Situations, Actions and Causal Laws') in {\it
  Semantic Information Processing}, ed. M. Minsky (Cambridge, MA: MIT Press
  (1963)).

\bibitem{mccarthy:conscious}
McCarthy, J.:
\newblock Making robots conscious.
\newblock In Furukawa, K., Michie, D., Muggleton, S., eds.: Machine
  Intelligence 15: Intelligent Agents.
\newblock Oxford University Press, Oxford (1996)

\bibitem{Horn:1951aa}
Horn, A.:
\newblock On sentences which are true of direct unions of algebras.
\newblock Journal of Symbolic Logic \textbf{16}(1) (1951)  14--21

\bibitem{McCarthy:1969aa}
McCarthy, J., Hayes, P.J.:
\newblock Some philosophical problems from the standpoint of artificial
  intelligence.
\newblock Machine Intelligence \textbf{4} (1969)

\bibitem{Maturana:1987aa}
Maturana, H.R., Varela, F.J.:
\newblock Tree of Knowledge: Biological Roots of Human Understanding.
\newblock Shambala (1987)

\bibitem{Braitenberg:1986aa}
Braitenberg, V.:
\newblock Vehicles: Experiments in Synthetic Psychology.
\newblock The MIT Press (1986)

\bibitem{Nilsson:1970aa}
Nilsson, N.J., Fikes, R.E.:
\newblock Strips: A new approach to the application of theorem proving to
  problem solving.
\newblock Technical Report~43, Stanford Research Institute, SRI, Menlo Park, CA
  (October 1970)

\bibitem{Muller:2008ab}
M{\"u}ller, M.E., Krebs, F., Hielscher, F.:
\newblock {Relational Cognitive Structures for Intelligent Agent and Robot
  Control}.
\newblock In: Systems, Man, and Cybernetics (SMC-2008), IEEE (2008)

\bibitem{Bun:87}
Buntine, W.:
\newblock Induction of {H}orn-clauses: methods and the plausible generalization
  algorithm.
\newblock \textbf{26} (1987)  499--520

\bibitem{mug:cigol1}
Muggleton, S., Buntine, W.:
\newblock Towards constructive induction in first-order predicate calculus.
\newblock TIRM 88-03, The Turing Institute, Glasgow (1988)

\bibitem{mugg:der}
Muggleton, S., Raedt, L.D.:
\newblock Inductive logic programming: Theory and methods.
\newblock Journal of Logic Programming  629--679

\bibitem{Raedt:2008aa}
Raedt, L.D.:
\newblock Logical and Relational Learning.
\newblock Cognitive Technologies. Springer (2008)

\bibitem{Scott89learningnovel}
Scott, P.D., Markovitch, S.:
\newblock Learning novel domains through curiosity and conjecture.
\newblock In: In Proceedings of International Joint Conference for Artificial
  Intelligence, Morgan Kaufmann (1989)  669--674

\bibitem{Shen92complementarydiscrimination}
Shen, W.M.:
\newblock Complementary discrimination learning with decision lists.
\newblock In: In Proceedings Tenth National Conference on Artificial
  Intelligence (pp. 153-158). Menlo Park, Ca, AAAI Press (1992)  153--158

\bibitem{Shen94discoveryas}
Shen, W.M.:
\newblock Discovery as autonomous learning from the environment.
\newblock In: Machine Learning, Computer Science Press (1994)  143--165

\bibitem{Leban:2008:ERD:1431080.1431092}
Leban, G., \v{Z}abkar, J., Bratko, I.:
\newblock An experiment in robot discovery with ilp.
\newblock In: Proceedings of the 18th international conference on Inductive
  Logic Programming. ILP '08, Berlin, Heidelberg, Springer-Verlag (2008)
  77--90

\bibitem{Aleph}
Srinivasan, A.:
\newblock The Aleph Manual.
\newblock University of Oxford. (2007)

\bibitem{muggsrin:progol}
Muggleton, S.:
\newblock Inverting entailment and {Progol}.
\newblock In Furukawa, K., Michie, D., Muggleton, S., eds.: Machine
  Intelligence 14.
\newblock Oxford University Press (1995)

\end{thebibliography}

\end{document}